\documentclass[runningheads,a4paper]{llncs}
\usepackage{dsfont}
\usepackage{amsmath}
\usepackage{amssymb}
\usepackage{graphicx}
\usepackage{subfigure}
\usepackage{wrapfig}
\usepackage{cite}
\usepackage{bm}
\usepackage{hyperref}
\usepackage{url}

\DeclareMathOperator*{\argmin}{arg\,min}

\begin{document}

\mainmatter  

\title{Estimation of Tissue Microstructure Using a Deep Network Inspired by a Sparse Reconstruction Framework}

\titlerunning{Estimation of Tissue Microstructure Using a Deep Network}

%
%
\author{Chuyang Ye$^{1,2}$}
\authorrunning{C. Ye}

\institute{$^{1}$Brainnetome Center \& $^{2}$National Laboratory of Pattern Recognition, \\Institute of Automation, Chinese Academy of Sciences, Beijing, China}

%
%

\toctitle{}
\tocauthor{}
\maketitle

\begin{abstract}
\textit{Diffusion magnetic resonance imaging} (dMRI) provides a unique tool for noninvasively probing the microstructure of the neuronal tissue. 
The NODDI model has been a popular approach to the estimation of tissue microstructure in many neuroscience studies. It represents the diffusion signals with three types of diffusion in tissue: intra-cellular, extra-cellular, and cerebrospinal fluid compartments.
However, the original NODDI method uses a computationally expensive procedure to fit the model and could require a large number of diffusion gradients for accurate microstructure estimation, which may be impractical for clinical use.
Therefore, efforts have been devoted to efficient and accurate NODDI microstructure estimation with a reduced number of diffusion gradients.
In this work, we propose a deep network based approach to the NODDI microstructure estimation, which is named \textit{Microstructure Estimation using a Deep Network}~(MEDN). 
Motivated by the AMICO algorithm which accelerates the computation of NODDI parameters, we formulate the microstructure estimation problem in a dictionary-based framework.
The proposed network comprises two cascaded stages.  
The first stage resembles the solution to a dictionary-based sparse reconstruction problem and the second stage computes the final microstructure using the output of the first stage. The weights in the two stages are jointly learned from training data, which is obtained from training dMRI scans with diffusion gradients that densely sample the $q$-space. 
The proposed method was applied to brain dMRI scans, where two shells each with 30 gradient directions (60 diffusion gradients in total) were used. Estimation accuracy with respect to the gold standard was measured and the results demonstrate that MEDN outperforms the competing algorithms.

\keywords{diffusion MRI, NODDI, microstructure, deep network}
\end{abstract}

\section{Introduction}
\label{sec:intro}

\textit{Diffusion magnetic resonance imaging} (dMRI) provides a unique tool for noninvasively probing the microstructure of the neuronal tissue by capturing the displacement pattern of water molecules~\cite{Johansen}. \textit{Diffusion tensor imaging} (DTI) was first developed to model the anisotropy of water diffusion using a Gaussian model, where fractional anisotropy and mean diffusivity can be computed to describe tissue microstructure. More complex methods have been proposed for improved diffusion modeling using biophysical models consisting of different tissue compartments, such as CHARMED~\cite{Assaf}, ActiveAx~\cite{Alexander}, and NODDI~\cite{Zhang}.

Among the existing algorithms for microstructure estimation, the NODDI model has been a popular choice in a number of scientific studies, for example, on brain development~\cite{Kelly} or pathological changes caused by diseases~\cite{Kamagata}. The NODDI model distinguishes three different types of diffusion in tissue, leading to the intra-cellular, extra-cellular, and \textit{cerebrospinal fluid} (CSF) compartments~\cite{Zhang}. By relating these compartments with observed diffusion signals, the parameters in the NODDI model are estimated with a maximum likelihood approach. The contribution of each compartment, the mean orientation of the intra-cellular compartment, and the orientation dispersion are then achieved, which give estimates of the tissue microstructural organization. 

The NODDI model uses a computationally expensive procedure to fit the model, and thus requires powerful computer clusters and/or takes a long computation time~\cite{Daducci2015}. To efficiently solve the NODDI model, a dictionary-based framework has been proposed in the AMICO algorithm~\cite{Daducci2015}, where the microstructure estimation is accelerated drastically. 
AMICO computes the mean orientation beforehand using DTI and estimates the CSF volume fraction, intra-cellular volume fraction, and orientation dispersion. First, it uses a dictionary that encodes discretized NODDI parameters to represent the diffusion signals. The \textit{mixture fractions}~(MFs) of the dictionary atoms can be estimated by solving a regularized least squares problem. The MF associated with the CSF atom provides an estimate of the CSF volume fraction. The other MFs, after normalization, linearly weight the discretized NODDI parameters to compute the rest of the microstructural properties.
However, NODDI or AMICO could require a large number of diffusion gradients for accurate microstructure estimation, which may limit their clinical use.
Thus, a multi-layer perceptron was used in~\cite{Golkov} to estimate scalar quantities including the NODDI parameters with a reduce number of diffusion gradients.

Efficient and accurate NODDI microstructure estimation using the number of diffusion gradients that is clinically practical (for example, around 60) is still an open problem.
In this work, we design a deep network to predict the NODDI microstructural properties.
The method is named \textit{Microstructure Estimation using a Deep Network}~(MEDN). Because the mean orientation can be estimated accurately using the simple DTI model~\cite{Daducci2015}, we focus on the scalar NODDI parameters like~\cite{Daducci2015}. 
Unlike~\cite{Golkov}, where a general multi-layer perceptron is used, the deep network in MEDN is designed specifically for the estimation of NODDI parameters.
The proposed network structure is motivated by the AMICO~\cite{Daducci2015} procedure and comprises two cascaded stages where all the weights are jointly learned for microstructure prediction.
The first stage uses a network structure that unfolds an iterative process similar to iterative hard thresholding~\cite{Blumensath}, and can solve a dictionary-based sparse reconstruction problem after the network weights are learned~\cite{Wang}. In the second stage, one of the output of the first stage corresponds to the CSF volume fraction; the other outputs are normalized and weighted to predict the intra-cellular volume fraction and orientation dispersion (after a transformation), where the weights are also learned. 
Like~\cite{Golkov}, for each dMRI dataset acquired with a fixed imaging protocol, one deep network is trained. To generate the training data, we use a strategy similar to~\cite{Golkov}, which requires training dMRI scans acquired with diffusion gradients that densely sample the $q$-space. The microstructure estimated by AMICO on the training images are then used to train the network, where  the sum of the mean squared errors of the CSF volume fraction, intra-cellular volume fraction, and orientation dispersion is used as the loss function.
The proposed method was evaluated on brain dMRI scans, where two shells each with 30 gradient directions were used, and the results demonstrate that MEDN outperforms the competitors.

\section{Methods}
\label{sec:method}

\subsection{Background: NODDI and AMICO for Tissue Microstructure Estimation}
\label{sec:bg}
NODDI models the neuronal tissue with three types of microstructural environments, which are the intra-cellular, extra-cellular, and CSF compartments~\cite{Zhang}. The water diffusion in each compartment has different distributions and thus different response functions to diffusion gradients. Suppose the number of diffusion gradients is $K$, the diffusion signal associated with the $k$-th ($k=1,\ldots,K$) diffusion gradient at a voxel is $S_{k}$, and the signal without diffusion weighting is $S_{0}$. The normalized signal $y_{k}=S_{k}/S_{0}$  is modeled using the three compartments:
\begin{eqnarray}
y_{k} = (1-v_{\mathrm{iso}})(v_{\mathrm{ic}}A_{\mathrm{ic},k} + (1-v_{\mathrm{ic}})A_{\mathrm{ec},k}) + v_{\mathrm{iso}}A_{\mathrm{iso},k},
\label{eqn:dwi}
\end{eqnarray}
where $A_{\mathrm{ic},k}$, $A_{\mathrm{ec},k}$, and $A_{\mathrm{iso},k}$ are the normalized signals of the intra-cellular, extra-cellular, and CSF compartments, respectively; $v_{\mathrm{ic}}$, $1-v_{\mathrm{ic}}$, and $v_{\mathrm{iso}}$ are the volume fractions of the intra-cellular, extra-cellular, and CSF compartments, respectively~\cite{Zhang}.

In NODDI, $A_{\mathrm{ic},k}$ is represented using a stick model, where the orientation $\bm{\mu}$ has a Watson distribution with a concentration parameter $\kappa$ that can measure the \textit{orientation dispersion}~(OD) by
\begin{eqnarray}
\mathrm{OD} = \frac{2}{\pi}\arctan(1/\kappa).
\label{eqn:od}
\end{eqnarray} 
$A_{\mathrm{ec},k}$ is modeled as anisotropic Gaussian diffusion, which is dependent on both $v_{\mathrm{ic}}$ and $\kappa$. $A_{\mathrm{iso},k}$ is modeled as isotropic Gaussian diffusion with predetermined diffusivity. For the specific design of the signal models for these compartments, we refer readers to~\cite{Zhang}.
Using the diffusion signals associated with all $K$ diffusion gradients, the parameters $v_{\mathrm{ic}}$, $\kappa$, $\bm{\mu}$, and $v_{\mathrm{iso}}$ are estimated with a maximum likelihood approach~\cite{Zhang}, and OD is derived from $\kappa$ according to Eq.~(\ref{eqn:od})

The original nonlinear approach to the NODDI model fitting in~\cite{Zhang} is very time-consuming and could cause practical problems for application to large cohorts~\cite{Daducci2015}.
Therefore, AMICO~\cite{Daducci2015}  is proposed to accelerate the estimation. 
It first decouples the estimation of mean orientations $\bm{\mu}$ and the other scalar quantities, and computes $\bm{\mu}$ using DTI. Then, the distinct water pools orientated in the direction $\bm{\mu}$ can be accounted for using a linear dictionary-based formulation
\begin{eqnarray}
\bm{y} = \mathbf{\Phi}_{\bm{\mu}} \bm{f} + \bm{\eta},
\label{eqn:noddi}
\end{eqnarray}
where $\bm{y}=(y_{1},\ldots,y_{K})^T$ is the observed signal vector, $\mathbf{\Phi}_{\bm{\mu}}$ is the dictionary, $\bm{f}$ is the MFs of the dictionary atoms,  and $\bm{\eta}$ is noise. The dictionary is computed from a fixed set of discretized $v_{\mathrm{ic}}$ and $\kappa$. It can be written as $\mathbf{\Phi}_{\bm{\mu}} = \left[\mathbf{\Phi}^{\mathrm{a}}_{\bm{\mu}}|\mathbf{\Phi}^{\mathrm{i}} \right]$, where $\mathbf{\Phi}^{\mathrm{a}}_{\bm{\mu}}\in \mathbb{R}^{K\times N_{\mathrm{a}}}$ comprises $N_{\mathrm{a}}$ columns of anisotropic signals of the coupled intra- and extra-cellular compartments corresponding to combinations of specific discretized $v_{\mathrm{ic}}$ and $\kappa$, and $\mathbf{\Phi}^{\mathrm{i}}\in \mathbb{R}^{K\times 1}$ comprises the signal terms of a constant isotropic diffusion.
The discretized $v_{\mathrm{ic}}$ and $\kappa$ associated with the $j$-th column in $\mathbf{\Phi}^{\mathrm{a}}_{\bm{\mu}}$ are denoted by $\tilde{v}_{\mathrm{ic},j}$ and $\tilde{\kappa}_{j}$, respectively.
AMICO uses 12 discretized $v_{\mathrm{ic}}$ and 12 discretized $\kappa$, leading to $N_{\mathrm{a}}=144$ combinations.
$\bm{f}$ can be denoted by $\bm{f} = (f_{1},\ldots,f_{N_{\mathrm{a}}},f_{N_{\mathrm{a}}+1})^T$, where $\bm{f}^{\mathrm{a}}=(f_{1},,\ldots,f_{N_{\mathrm{a}}})^T$ and $f^{\mathrm{i}}=f_{N_{\mathrm{a}}+1}$ are associated with $\mathbf{\Phi}^{\mathrm{a}}_{\bm{\mu}}$ and $\mathbf{\Phi}^{\mathrm{i}}$, respectively.

The MFs are estimated by solving
\begin{eqnarray}
\hat{\bm{f}}=\argmin_{\bm{f}\geq 0}||\mathbf{\Phi}_{\bm{\mu}}\bm{f}-\bm{y}||_{2}^{2} + \alpha||\bm{f}||_{2}^{2} + \beta||\bm{f}||_{1},
\label{eqn:amico}
\end{eqnarray}
where $\alpha$ and $\beta$ are weights specified by users. Then, the NODDI parameters are computed from the MFs and discretized $v_{\mathrm{ic}}$ and $\kappa$ as follows
\begin{eqnarray}
v_{\mathrm{ic}}=\frac{\sum_{j=1}^{N_{\mathrm{a}}}\tilde{v}_{\mathrm{ic},j}\hat{f}_{j}}{\sum_{j=1}^{N_{\mathrm{a}}}\hat{f}_{j}},
\kappa=\frac{\sum_{j=1}^{N_{\mathrm{a}}}\tilde{\kappa}_{j}\hat{f}_{j}}{\sum_{j=1}^{N_{\mathrm{a}}}\hat{f}_{j}},\,\mathrm{and}\,\,
v_{\mathrm{iso}} = \hat{f}_{N_{\mathrm{a}}+1},
\label{eqn:computation}
\end{eqnarray}
and OD is computed from $\kappa$ using Eq.~(\ref{eqn:od}).
It is demonstrated in~\cite{Daducci2015} that AMICO reduces the computational time by two orders of magnitude.

\subsection{Tissue Microstructure Estimation Using a Deep Network}

\begin{figure}[!t]
  \centering
	\includegraphics[width=0.85\columnwidth]{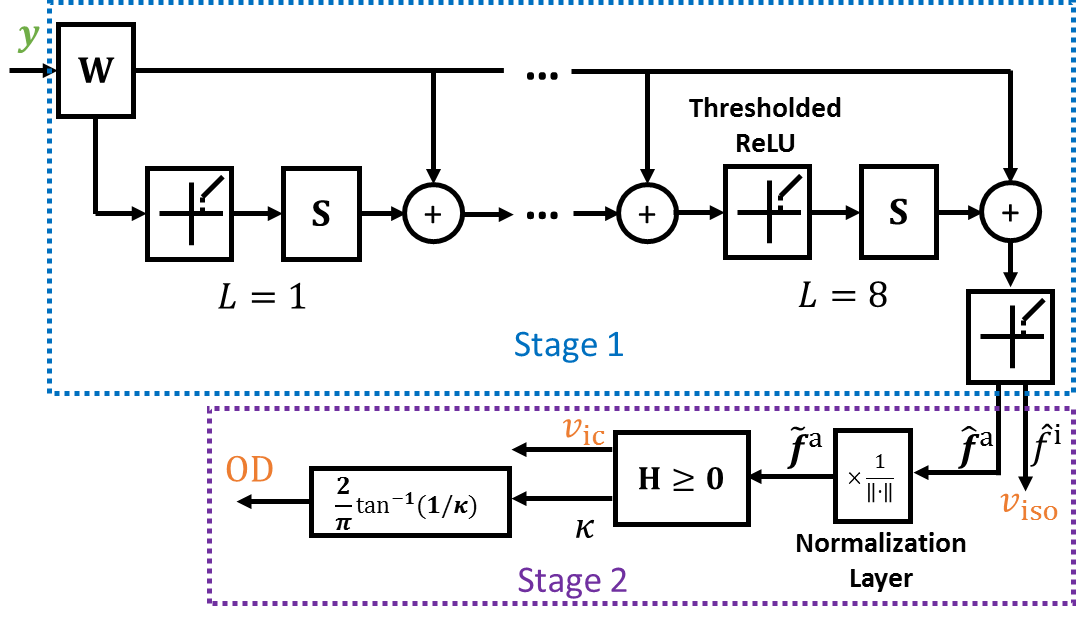}
\caption{The deep network designed for microstructure estimation. The input and output of the network are indicated by the green and orange colors, respectively.}
\label{fig:dn}
\end{figure}

The deep network has been applied to many computer vision tasks~\cite{LeCun}, and in this work we explore its use in tissue microstructure estimation.
The design of a deep network structure can be motivated in many ways, for example, by the organization of neurons in the brain~\cite{Krizhevsky} or by procedures performed by specific algorithms~\cite{Gregor,Wang}. 
Our design of the network structure for microstructure estimation belongs to the latter category.
As introduced in Sect.~\ref{sec:bg}, AMICO consists of two steps: 1) solving a regularized least squares problem (see Eq.~(\ref{eqn:amico})) and 2) computing the NODDI parameters using the MFs and discretized parameters (see Eq.~(\ref{eqn:computation})). 
Motivated by these steps, we construct a deep network whose structure is similar to the AMICO procedure, while the weights in the network are learned instead of predetermined. 
The proposed network structure is shown in Fig.~\ref{fig:dn}, which comprises two cascaded stages. The input and output of the network are indicated by the green and orange colors, respectively. The design is explained in detail in the following paragraphs.

\subsubsection{Stage One}The first stage takes the observed normalized diffusion signals $\bm{y}$ as input and seeks to solve a regularized least squares problem with learned parameters. We notice that in~\cite{Daducci2015} the weight $\alpha$ is much smaller than $\beta$ in Eq.~(\ref{eqn:amico}), and setting $\alpha=0$ can still achieve low estimation errors. Thus, we let $\alpha=0$ and Eq.~(\ref{eqn:amico}) becomes an $\ell_{1}$-norm regularized least squares problem, which is an approximation of the (nonnegative) sparse reconstruction problem
\begin{eqnarray}
\hat{\bm{f}} = \argmin_{\bm{f}\geq 0}||\mathbf{\Phi}\bm{f}-\bm{y}||_{2}^{2} + \beta||\bm{f}||_{0}.
\label{eqn:l0}
\end{eqnarray}
For convenience, here we have dropped the symbol $\bm{\mu}$. 
Conventionally, the sparse reconstruction problem can be solved using iterative hard thresholding (IHT)~\cite{Blumensath}, which iteratively updates the estimate. Specifically, at iteration $t+1$
\begin{eqnarray}
\bm{f}^{t+1} = h_{\lambda}(\mathbf{W}\bm{y}+\mathbf{S}\bm{f}^{t}),
\label{eqn:iht}
\end{eqnarray}
where $\mathbf{W}=\mathbf{\Phi}^{T}$, $\mathbf{S}=\mathbf{I}-\mathbf{\Phi}^{T}\mathbf{\Phi}$, and $h_{\lambda}(\cdot)$ is a thresholding operator with a parameter $\lambda>0$
\begin{eqnarray}
[h_{\lambda}(\bm{a})]_{i} = 
\begin{cases}
0\quad &\mathrm{if}\quad a_{i} < \lambda\\
a_{i} \quad &\mathrm{if}\quad a_{i} \geq \lambda
\end{cases}
.
\label{eqn:operator}
\end{eqnarray}
Note that due to the constraint $\bm{f}\geq \bm{0}$, $[h_{\lambda}(\bm{a})]_{i}$ is always zero when $a_{i}$ is negative.

Motivated by the iterative process in IHT, a feed-forward network structure can be constructed by unfolding and truncating this process~\cite{Wang}, which is indicated by the blue box in Fig.~\ref{fig:dn}. 
The number of layers in this stage is eight, which lies in the range of the numbers used by previous works~\cite{Wang,Xin,Sprechmann}. 
We assume $\bm{f}^{0}=0$ and the thresholded \textit{rectified linear unit} (ReLU)~\cite{Konda} corresponds to the operator $h_{\lambda}(\cdot)$ ($\lambda=0.01$ in this work). 
The update of $\bm{f}$ according to Eq.~(\ref{eqn:iht}) is completed after each thresholded ReLU.
Here, instead of using $\mathbf{W}$ and $\mathbf{S}$ predetermined by $\mathbf{\Phi}$, $\mathbf{W}\in\mathbb{R}^{N\times K}$ and $\mathbf{S}\in \mathbb{R}^{N\times N}$ in the network are learned from training data, and the dimension $N$ is to be specified by the users. Greater $N$ leads to more weights to be learned in the network, and in this work we empirically set $N=301$. Note that $\mathbf{S}$ is shared among layers, thus increasing the number of layers does not increase the number of weights to be learned. 

It was demonstrated in~\cite{Xin} that learned layer-wise fixed weights could guarantee successful sparse reconstruction across a wider range of \textit{restricted isometry property}~(RIP) conditions than IHT. In addition, because usually we only seek to solve a problem where inputs are similar to the training data, the problem is smaller than a general sparse reconstruction problem for all possible inputs, and it is possible to use learned weights to achieve superior reconstruction~\cite{Gregor}.

Note that in the original AMICO framework, different $\mathbf{\Phi}_{\bm{\mu}}$ is needed for different $\bm{\mu}$. However, in this work we only construct one deep network for microstructure estimation for all possible $\bm{\mu}$, and it can be interpreted in the following way.
The mean orientations can be discretized as well~\cite{Landman,FORNI}, which gives a basis orientation set $\mathcal{U}=\{\tilde{\bm{\mu}}_{i}\}_{i=1}^{|\mathcal{U}|}$ ($|\mathcal{U}|$ is the cardinality of $\mathcal{U}$). Then, the dictionary matrix can be expanded to include the signal terms associated with the discretized $\bm{\mu}$, $v_{\mathrm{ic}}$, and $\kappa$, so that $\mathbf{\Phi} = \left[\mathbf{\Phi}^{\mathrm{a}}_{\tilde{\bm{\mu}}_{1}}|\ldots|\mathbf{\Phi}^{\mathrm{a}}_{\tilde{\bm{\mu}}_{|\mathcal{U}|}}|\mathbf{\Phi}^{\mathrm{i}} \right]$. The microstructure can still be computed from the MFs associated with $\mathbf{\Phi}$ using Eq.~(\ref{eqn:computation}).

\subsubsection{Stage Two}The second stage (indicated by the purple box in Fig.~\ref{fig:dn}) computes the NODDI parameters using the output of the first stage. $v_{\mathrm{iso}}$ is immediately achieved from the entry $\hat{f}^{\mathrm{i}}$ in $\hat{\bm{f}}$ that corresponds to the CSF compartment. From Eq.~(\ref{eqn:computation}), we see that $v_{\mathrm{ic}}$ and $\kappa$ are computed by linearly transforming the normalized MFs of anisotropic diffusion compartments. Thus, the other entries $\hat{\bm{f}}^{\mathrm{a}}$ in $\hat{\bm{f}}$ are first normalized by the normalization layer in the second stage. Note that to ensure numerical stability, we use $\tilde{\bm{f}}^{\mathrm{a}} = (\hat{\bm{f}}^{\mathrm{a}}+\tau\bm{1})/||\hat{\bm{f}}^{\mathrm{a}}+\tau\bm{1}||_{1}$ for the normalization, where $\tau=10^{-10}$.
Then, the computation of $v_{\mathrm{ic}}$ and $\kappa$ resembling Eq.~(\ref{eqn:computation}) can be written in the matrix form
\begin{eqnarray}
\left[v_{\mathrm{ic}} \,\, \kappa\right]^{T} = \mathbf{H}\tilde{\bm{f}}^{\mathrm{a}}, 
\label{eqn:computation2}
\end{eqnarray}
where $\mathbf{H}\in\mathbb{R}^{2\times (N-1)}$ contains the weights of $\tilde{\bm{f}}^{\mathrm{a}}$ and is to be learned. Because the discretized $v_{\mathrm{ic}}$ and $\kappa$ in Eq.~(\ref{eqn:computation}) are nonnegative, here we require each element in $\mathbf{H}$ to be nonnegative. OD is computed from $\kappa$ using Eq.~(\ref{eqn:od}). The final outputs of the network are $v_{\mathrm{ic}}$, $v_{\mathrm{iso}}$, and OD.

\subsection{Training and Evaluation}

The weights in the two stages are learned jointly.
We use the sum of the mean squared errors of $v_{\mathrm{ic}}$, $v_{\mathrm{iso}}$, and OD as the loss function and the Adam algorithm~\cite{Kingma} as the optimizer, where the learning rate is 0.0001, the batch size is 128, and the number of epochs is 10. 
Similar to~\cite{Golkov}, 10\% of the training samples were used as a validation set to prevent overfitting. The network is implemented using Keras\footnote{\url{http://keras.io/}}.

We use a training strategy similar to~\cite{Golkov}. 
Because the observed diffusion signals are dependent on the diffusion gradients used in the imaging protocol, for each dataset of dMRI scans that are acquired with a fixed set $\mathcal{G}$ of diffusion gradients, one deep network needs to be trained for microstructure prediction.
Because ground truth microstructure is difficult to acquire, to generate training samples, training dMRI scans should be acquired with a set $\tilde{\mathcal{G}}$ of diffusion gradients that densely sample the $q$-space, where $\mathcal{G}\subseteq\tilde{\mathcal{G}}$. Each voxel in the training images represents a training sample, and the microstructure computed by \hyphenation{AMICO}AMICO on the training dMRI scans using all diffusion gradients $\tilde{\mathcal{G}}$ and the diffusion signals associated with $\mathcal{G}$ were used to train the network. 

Using the trained network, tissue microstructure at each voxel on a test image can be estimated. To quantitatively evaluate the estimation performance, the gold standard should be obtained, with which estimation results are compared. Similar to the generation of training images, for each test dMRI scan with diffusion gradients $\mathcal{G}$, diffusion gradients $\tilde{\mathcal{G}}$ densely sampling the $q$-space were also applied to compute the gold standard of tissue microstructure using AMICO. Note that $\tilde{\mathcal{G}}$ was only used for computing the gold standard for evaluation, and was not used in the test phase. The mean absolute difference was used to compute the disagreement between the estimates and the gold standard.

\section{Results}
\label{sec:exp}

\begin{figure}[!t]
  \centering
	\includegraphics[width=0.97\columnwidth]{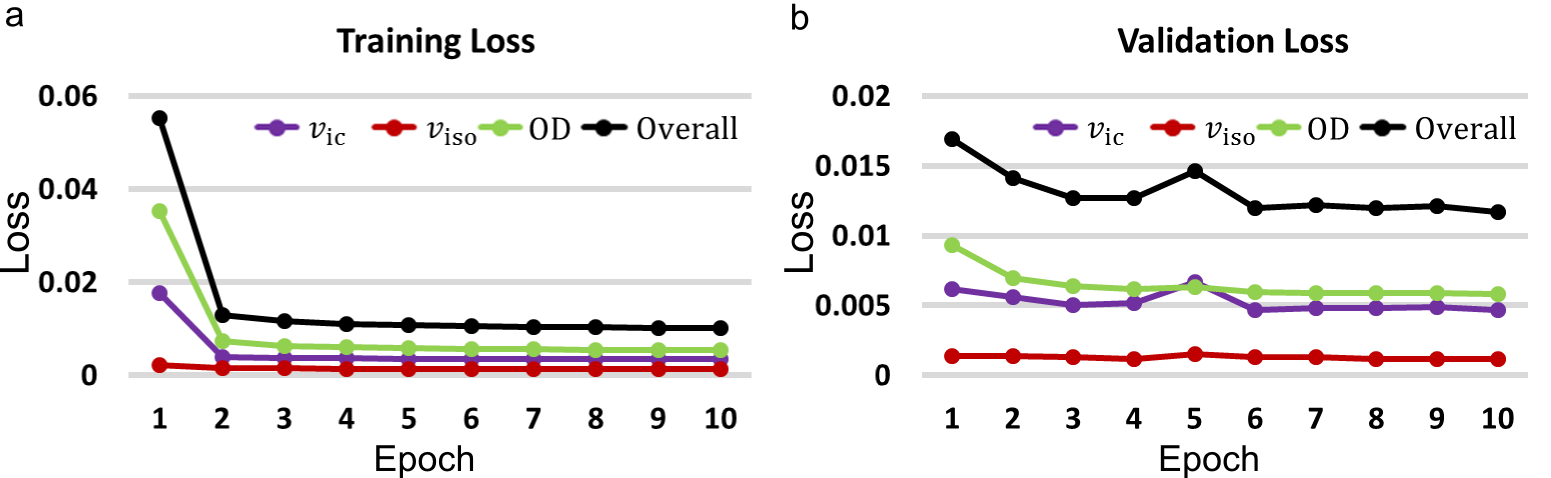}
\caption{The training and validation loss after each epoch in the training phase.}
\label{fig:training}
\end{figure}

\begin{figure}[!t]
  \centering
	\includegraphics[width=0.97\columnwidth]{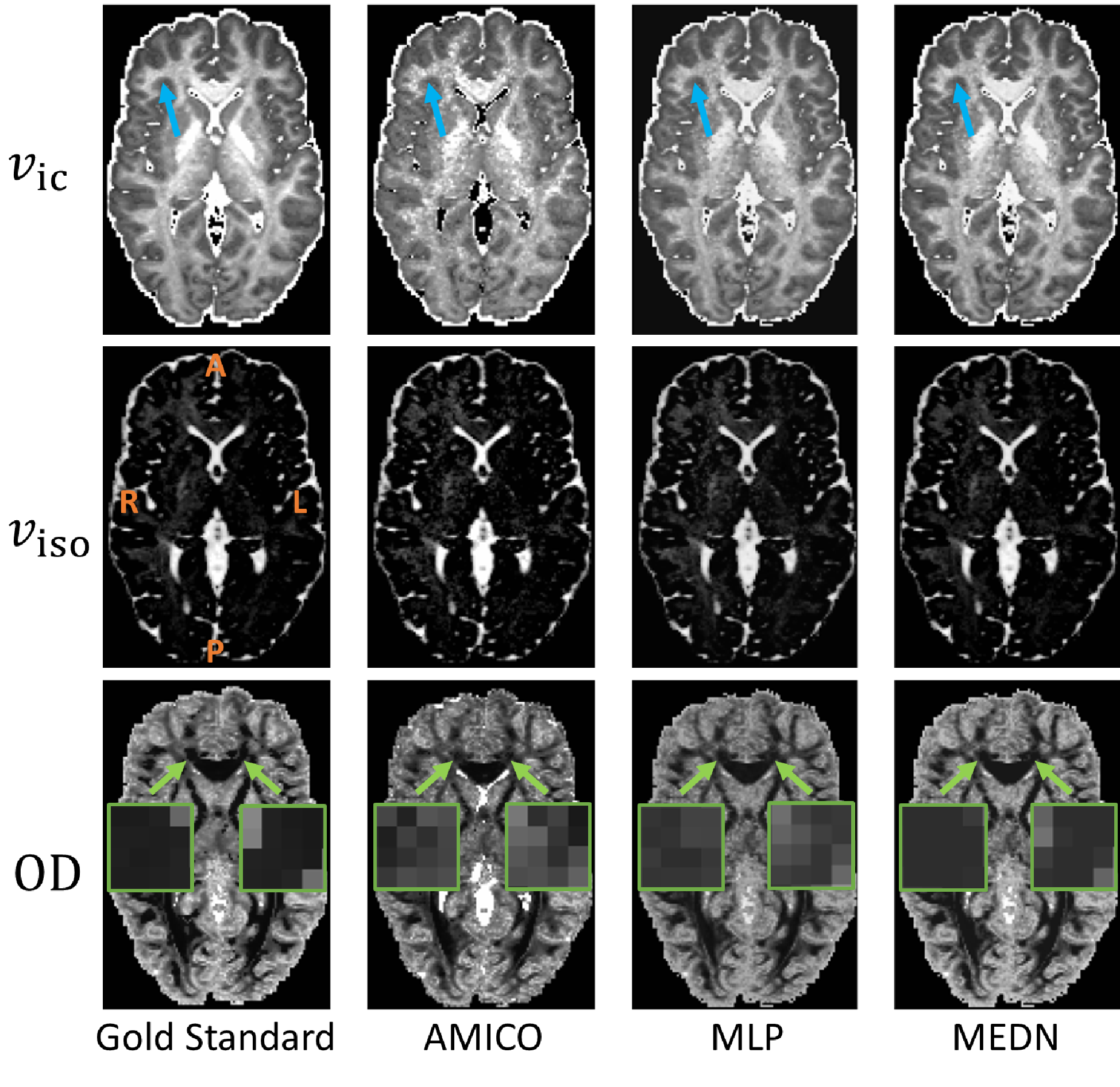}
\caption{Microstructure estimation of AMICO, MLP, and MEDN shown together with the gold standard on a representative subject. The orientation is indicated on the gold standard $v_{\mathrm{iso}}$ map (the left image in the middle row). The contrasts in the zoomed regions on the OD maps are enhanced with the same mapping.}
\label{fig:brain}
\end{figure}

The proposed method was applied to brain dMRI for evaluation on a 16-core Linux machine.
We randomly selected ten subjects from the Human Connectome Project~(HCP) dataset~\cite{VanEssen}.
The \textit{diffusion weighted images}~(DWIs) were acquired on a 3T MR scanner (ConnectomS, Siemens, Erlangen, Germany), where 270 diffusion gradients over three shells with $b$-values of 1000, 2000, and 3000~$\mathrm{s}/\mathrm{mm}^{2}$ were used. The resolution of the DWIs is 1.25~mm isotropic. Five subjects were randomly selected as training scans and the other five were used as test scans. 
For each training or test scan, 60 fixed diffusion gradients were selected as the diffusion gradients $\mathcal{G}$, and the normalized diffusion signals associated with $\mathcal{G}$ are the input to the network in the training or test phase, respectively. These 60 diffusion gradients resemble clinically achievable protocols. They consist of 30 gradient directions on each of the shell $b=1000,2000~\mathrm{s}/\mathrm{mm}^{2}$, and the gradient directions are approximately evenly distributed over the unit sphere.
The full set of 270 diffusion gradients were used to compute the training and gold standard microstructure for the training and test scans, respectively. 

The training process using the five training subjects took about 8.5 hours. The overall training loss and validation loss in the training phase are shown in Fig.~\ref{fig:training}, together with the loss of each microstructure quantity. We can see that both the training loss and validation loss become stable after ten epochs with the selected parameters of the network.

The trained network was then applied to the test dMRI scans for microstructure prediction. The estimation accuracy of MEDN was compared with that of AMICO (using the implementation and default parameters provided at \url{https://github.com/daducci/AMICO/}) and the deep network structure proposed in~\cite{Golkov}. The authors of~\cite{Golkov} used a \textit{multi-layer perceptron}~(MLP) to predict scalar quantities including NODDI parameters. The MLP consists of three hidden layers, each comprising 150 hidden units with a ReLU~\cite{Nair} activation function, and the dropout fraction is 0.1; 10\% of the training voxels were used as a validation set in the training phase.
In our experiments, the weights in the MLP were learned from the same training samples used by MEDN. The input of AMICO and MLP for the test scans is the same as that of MEDN.
For each test subject, both MLP and MEDN took about 30 minutes, and AMICO took about five hours. 

\begin{figure}[!t]
  \centering
	\includegraphics[width=0.97\columnwidth]{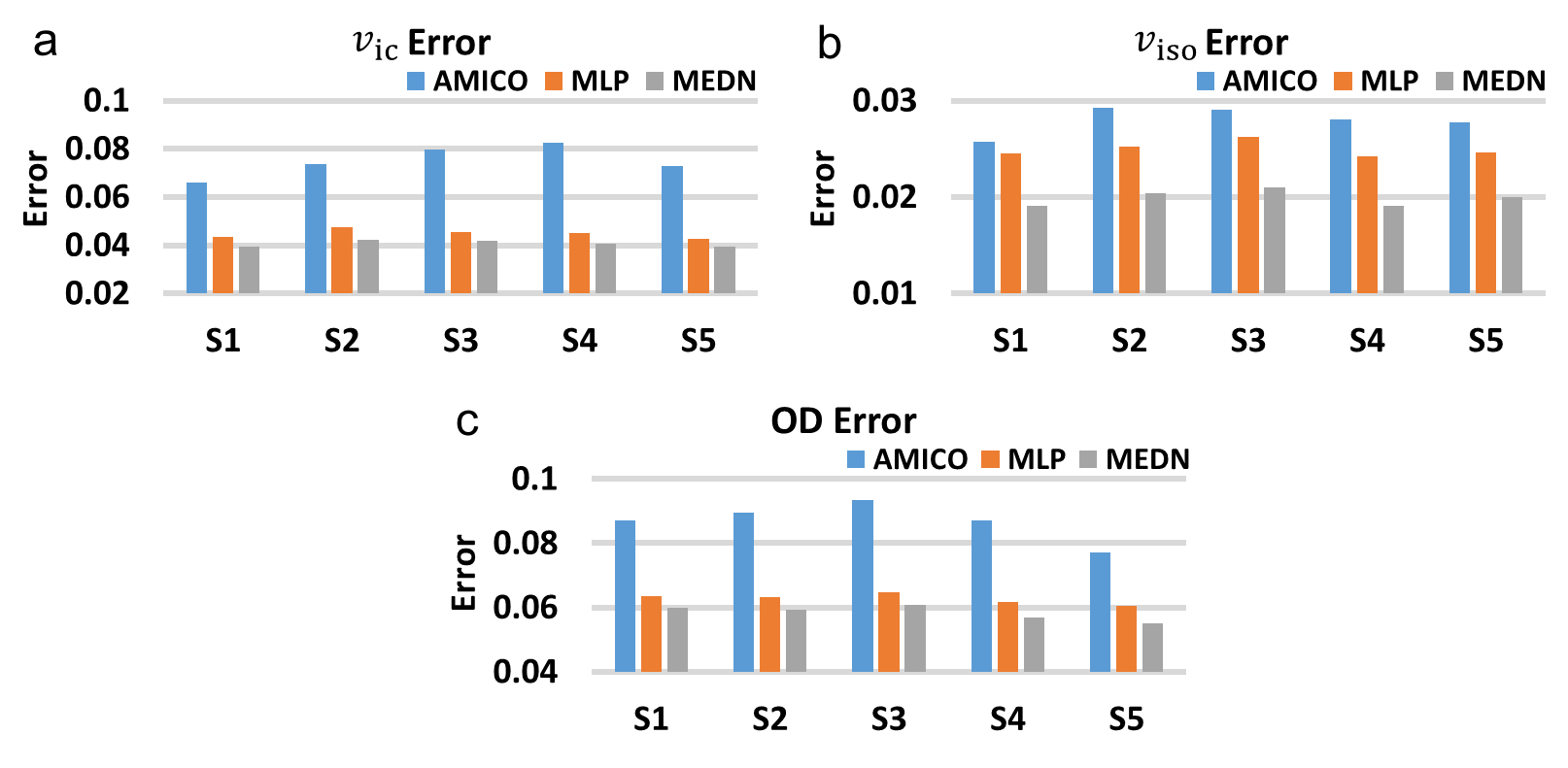}
\caption{Average errors of tissue microstructure estimation in the brain for each test subject (S1--S5): (a) $v_{\mathrm{ic}}$, (b) $v_{\mathrm{iso}}$, and (c) OD.}
\label{fig:errors}
\end{figure}

\begin{figure}[!t]
  \centering
	\includegraphics[width=0.97\columnwidth]{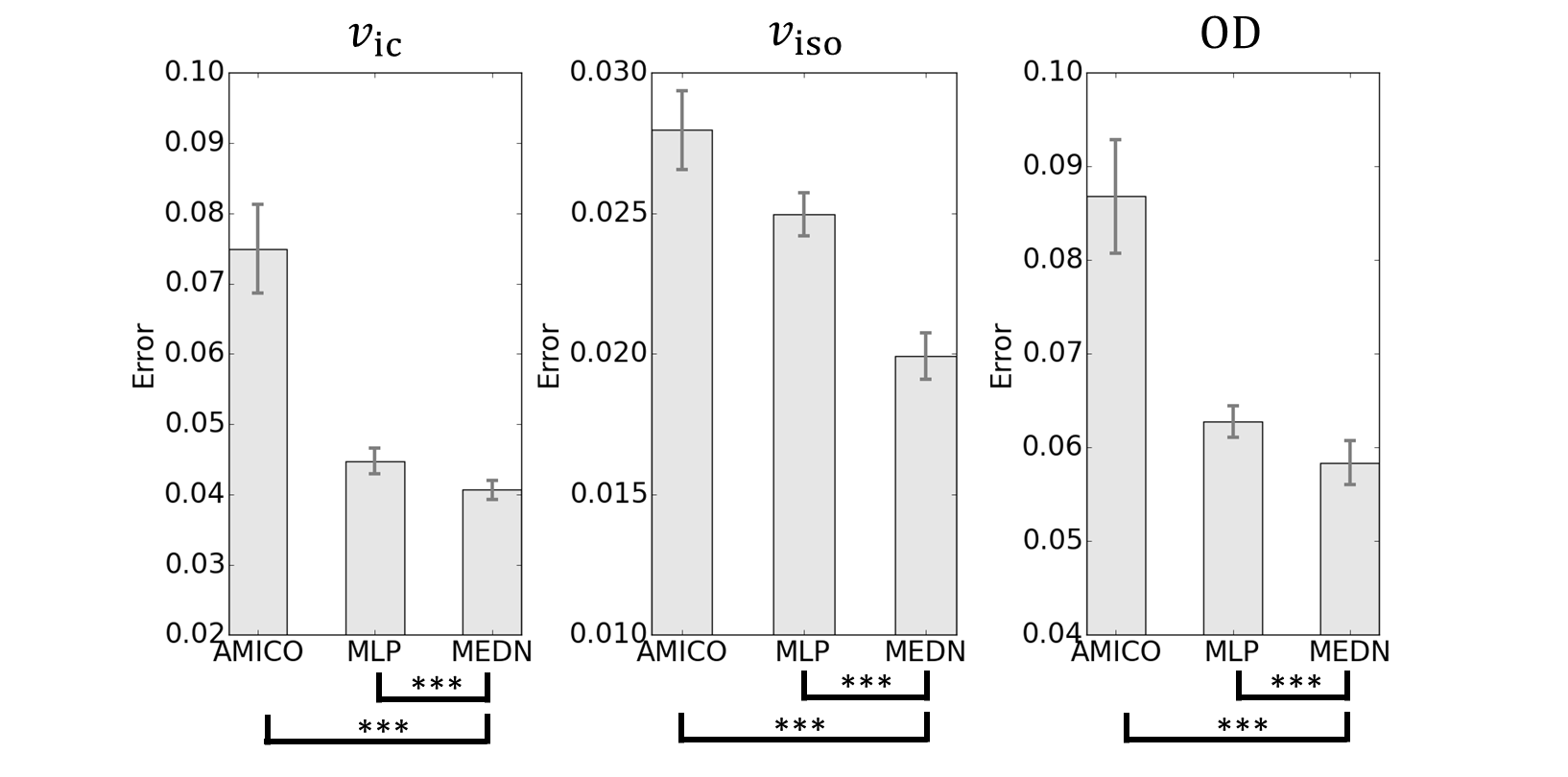}
\caption{Means and standard deviations of the average estimation errors in Fig.~\ref{fig:errors} computed using all five test subjects. MEDN was compared with AMICO and MLP using a paired Student's $t$-test, and asterisks indicate that the difference is significant ($^{***}p<0.001$).}
\label{fig:errors_all}
\end{figure}

Cross-sectional slices of the gold standard and estimated $v_{\mathrm{ic}}$, $v_{\mathrm{iso}}$, and OD on a representative subject are shown in Fig.~\ref{fig:brain} for qualitatively evaluation. Both MLP and MEDN produced a smoother $v_{\mathrm{ic}}$ map than AMICO (for example, see the regions pointed by the blue arrow), which better agrees with the gold standard. In the anterior corpus callosum (for example, the zoomed regions indicated by the green arrows) on the OD maps, the MEDN result is less noisy and better resembles the gold standard than the AMICO and MLP results.

The average microstructure estimation errors in the brain with respect to the gold standard are shown for AMICO, MLP, and MEDN for all five test subjects in Fig.~\ref{fig:errors}. In all cases MEDN achieves the lowest error. The means and standard deviations of the average estimation errors computed using the five test subjects are shown in Fig.~\ref{fig:errors_all}, and AMICO and MLP were compared with MEDN using a paired Student's $t$-test. The estimation error of MEDN is significantly smaller than that of AMICO and MLP for $v_{\mathrm{ic}}$, $v_{\mathrm{iso}}$, and OD. 
Compared with MLP which has the second best performance, MEDN reduces the mean errors of $v_{\mathrm{ic}}$, $v_{\mathrm{iso}}$, and OD by about 9\%, 20\%, and 7\%, respectively.

\section{Discussion}
\label{sec:discussion}

The original NODDI model fitting approach in~\cite{Zhang} can be very time-consuming. 
As reported in~\cite{Daducci2015}, the original NODDI computation took around 65 hours on a dMRI scan at a resolution much lower than that of the HCP data used in this work. Thus, performing the original NODDI model fitting would be prohibitive for the experiments in this work. 
AMICO has been shown to produce results comparable to~\cite{Zhang} and reduce the computational time by two orders of magnitude. 
Therefore, we used AMICO to compute the training data and gold standard of tissue microstructure.

The NODDI model can be improved by modeling anisotropic orientation dispersion that is widespread due to fiber bending and fanning~\cite{Tariq}. In~\cite{Tariq} the Bingham-NODDI model was proposed, which replaces the Watson distribution in NODDI with the Bingham distribution to allow estimation of anisotropic dispersion and thus introduces extra parameters.
It is possible to formulate the Bingham-NODDI model in a dictionary-based framework like AMICO, where the dictionary atoms are computed using the Bingham distribution with its discretized parameters. Thus, the first stage in MEDN still applies, and the second stage needs to be adapted to compute additional microstructure descriptors modeled by Bingham-NODDI.

NODDI and AMICO only assume one fiber orientation in a voxel while brain regions can contain crossing fibers~\cite{Auria2015}. Thus, AMICOx was proposed in~\cite{Auria2015}, which improves AMICO by expanding the dictionary to encode atoms corresponding to multiple precomputed fiber orientations. Since AMICOx also relies on solving a dictionary-based regularized least squares problem, we can still use the structure in the first stage in MEDN and adapt the second stage to compute orientation-specific microstructure in regions containing crossing tracts. 

The parameters in the network were empirically determined, such as $\lambda$ in the activation function $h_{\lambda}(\cdot)$ and the number $N$ of rows in $\mathbf{W}$ and $\mathbf{S}$. The results demonstrate that the selected parameters produce reasonable microstructure estimation. A thorough investigation of the impact of these parameters will be performed in the future. In addition, it is possible to learn the parameter $\lambda$ as well, where $\lambda$ can be encoded in two different multiplication modules before and after the thresholded ReLU instead of in the thresholded ReLU~\cite{Wang}.

\section{Conclusion}
\label{sec:conclusion}

We have proposed a deep network based approach, MEDN, to the prediction of tissue microstructure based on the NODDI model. 
MEDN comprises two stages, where the weights are learned jointly. The first stage resembles the solution to a sparse reconstruction problem and the second stage computes the microstructure using the output of the first stage. Results on brain dMRI data demonstrate that MEDN outperforms the competing methods.

\section*{Acknowledgement}

This work is supported by NSFC 61601461.
Data were provided by the Human Connectome Project, WU-Minn Consortium (Principal Investigators: David Van Essen and Kamil Ugurbil; 1U54MH091657).

\bibliographystyle{splncs03}
\bibliography{refs}

\begin{thebibliography}{10}
\providecommand{\url}[1]{\texttt{#1}}
\providecommand{\urlprefix}{URL }

\bibitem{Alexander}
Alexander, D.C., Hubbard, P.L., Hall, M.G., Moore, E.A., Ptito, M., Parker,
  G.J., Dyrby, T.B.: Orientationally invariant indices of axon diameter and
  density from diffusion {MRI}. NeuroImage  52(4),  1374--1389 (2010)

\bibitem{Assaf}
Assaf, Y., Basser, P.J.: Composite hindered and restricted model of diffusion
  {(CHARMED) MR} imaging of the human brain. NeuroImage  27(1),  48--58 (2005)

\bibitem{Auria2015}
Aur{\'\i}a, A., Romascano, D.P.R., Canales-Rodriguez, E., Wiaux, Y., Dirby,
  T.B., Alexander, D., Thiran, J.P., Daducci, A.: Accelerated microstructure
  imaging via convex optimisation for regions with multiple fibres ({AMICOx}).
  In: IEEE International Conference on Image Processing 2015. pp. 1673--1676.
  IEEE (2015)

\bibitem{Blumensath}
Blumensath, T., Davies, M.E.: Iterative thresholding for sparse approximations.
  Journal of Fourier Analysis and Applications  14(5-6),  629--654 (2008)

\bibitem{Daducci2015}
Daducci, A., Canales-Rodr{\'\i}guez, E.J., Zhang, H., Dyrby, T.B., Alexander,
  D.C., Thiran, J.P.: {Accelerated Microstructure Imaging via Convex
  Optimization (AMICO}) from diffusion {MRI} data. NeuroImage  105,  32--44
  (2015)

\bibitem{Golkov}
Golkov, V., Dosovitskiy, A., Sperl, J.I., Menzel, M.I., Czisch, M., S{\"a}mann,
  P., Brox, T., Cremers, D.: q-space deep learning: Twelve-fold shorter and
  model-free diffusion {MRI} scans. IEEE Transactions on Medical Imaging
  35(5),  1344--1351 (2016)

\bibitem{Gregor}
Gregor, K., LeCun, Y.: Learning fast approximations of sparse coding. In:
  International Conference on Machine Learning. pp. 399--406 (2010)

\bibitem{Johansen}
Johansen-Berg, H., Behrens, T.E.J.: Diffusion {MRI}: from quantitative
  measurement to in vivo neuroanatomy. Waltham: Academic Press (2013)

\bibitem{Kamagata}
Kamagata, K., Hatano, T., Okuzumi, A., Motoi, Y., Abe, O., Shimoji, K., Kamiya,
  K., Suzuki, M., Hori, M., Kumamaru, K.K., Hattori, N., Aoki, S.: Neurite
  orientation dispersion and density imaging in the substantia nigra in
  idiopathic {Parkinson} disease. European Radiology  26(8),  2567--2577 (2016)

\bibitem{Kelly}
Kelly, C.E., Thompson, D.K., Chen, J., Leemans, A., Adamson, C.L., Inder, T.E.,
  Cheong, J.L., Doyle, L.W., Anderson, P.J.: Axon density and axon orientation
  dispersion in children born preterm. Human Brain Mapping  37(9),  3080--3102
  (2016)

\bibitem{Kingma}
Kingma, D., Ba, J.: Adam: A method for stochastic optimization. arXiv preprint
  arXiv:1412.6980  (2014)

\bibitem{Konda}
Konda, K., Memisevic, R., Krueger, D.: Zero-bias autoencoders and the benefits
  of co-adapting features. arXiv preprint arXiv:1402.3337  (2014)

\bibitem{Krizhevsky}
Krizhevsky, A., Sutskever, I., Hinton, G.E.: {ImageNet} classification with
  deep convolutional neural networks. In: Advances in Neural Information
  Processing Systems. pp. 1097--1105 (2012)

\bibitem{Landman}
Landman, B.A., Bogovic, J.A., Wan, H., ElShahaby, F.E.Z., Bazin, P.L., Prince,
  J.L.: Resolution of crossing fibers with constrained compressed sensing using
  diffusion tensor {MRI}. NeuroImage  59(3),  2175--2186 (2012)

\bibitem{LeCun}
LeCun, Y., Bengio, Y., Hinton, G.: Deep learning. Nature  521(7553),  436--444
  (2015)

\bibitem{Nair}
Nair, V., Hinton, G.E.: Rectified linear units improve restricted boltzmann
  machines. In: Proceedings of the 27th International Conference on Machine
  Learning. pp. 807--814 (2010)

\bibitem{Sprechmann}
Sprechmann, P., Bronstein, A.M., Sapiro, G.: Learning efficient sparse and low
  rank models. IEEE Transactions on Pattern Analysis and Machine Intelligence
  37(9),  1821--1833 (2015)

\bibitem{Tariq}
Tariq, M., Schneider, T., Alexander, D.C., Wheeler-Kingshott, C.A.G., Zhang,
  H.: Bingham--{NODDI}: Mapping anisotropic orientation dispersion of neurites
  using diffusion {MRI}. NeuroImage  133,  207--223 (2016)

\bibitem{VanEssen}
Van~Essen, D.C., Smith, S.M., Barch, D.M., Behrens, T.E.J., Yacoub, E.,
  Ugurbil, K.: The {WU-Minn} human connectome project: An overview. NeuroImage
  80(0),  62--79 (2013)

\bibitem{Wang}
Wang, Z., Ling, Q., Huang, T.S.: Learning deep $\ell_0$ encoders. In: AAAI
  Conference on Artificial Intelligence. pp. 2194--2200 (2016)

\bibitem{Xin}
Xin, B., Wang, Y., Gao, W., Wipf, D.: Maximal sparsity with deep networks?
  arXiv preprint arXiv:1605.01636  (2016)

\bibitem{FORNI}
Ye, C., Zhuo, J., Gullapalli, R.P., Prince, J.L.: Estimation of fiber
  orientations using neighborhood information. Medical Image Analysis  32,
  243--256 (2016)

\bibitem{Zhang}
Zhang, H., Schneider, T., Wheeler-Kingshott, C.A., Alexander, D.C.: {NODDI}:
  Practical in vivo neurite orientation dispersion and density imaging of the
  human brain. NeuroImage  61(4),  1000--1016 (2012)

\end{thebibliography}
\end{document}